\documentclass[times,final]{elsarticle}

\usepackage{framed,multirow}

\usepackage{amssymb}
\usepackage{latexsym}
\usepackage{amsmath} 
\usepackage{booktabs}
\usepackage{url}
\usepackage{xcolor}
\usepackage{graphicx}
\usepackage{subfigure}
\usepackage[colorlinks,
linkcolor=blue,
anchorcolor=blue,
citecolor=blue]{hyperref} 
\usepackage{algorithm}
\usepackage{algorithmicx}
\usepackage{algpseudocode}
\usepackage{bm}
\usepackage{diagbox}
\definecolor{newcolor}{rgb}{.8,.349,.1}

\begin{document}


\begin{frontmatter}

\title{RBF-MGN:Solving spatiotemporal PDEs with Physics-informed Graph Neural Network }%
\tnotetext[tnote1]{This is an example for title footnote coding.}

\author[1]{Zixue Xiang}
\ead{xiangzixuebit@163.com}
\author[2]{Wei Peng}
\author[2]{Wen Yao \corref{cor1}}
\cortext[cor1]{Corresponding author}

\address[1]{College of Aerospace Science and Engineering, National University of Defense Technology, No. 109, Deya Road, Changsha 410073, China}
\address[2]{National Innovation Institute of Defense Technology, Chinese Academy of Military Science, No. 55, Fengtai East Street, Beijing 100071, China}


\begin{abstract}
Physics-informed neural networks (PINNs) have lately received significant attention as a representative deep learning-based technique for solving partial differential equations (PDEs). Most fully connected network-based PINNs use automatic differentiation to construct loss functions that suffer from slow convergence and difficult boundary enforcement. In addition, although convolutional neural network (CNN)-based PINNs can significantly improve training efficiency, CNNs have difficulty in dealing with irregular geometries with unstructured meshes. Therefore, we propose a novel framework based on graph neural networks (GNNs) and radial basis function finite difference (RBF-FD). We introduce GNNs into physics-informed learning to better handle irregular domains with unstructured meshes. RBF-FD is used to construct a high-precision difference format of the differential equations to guide model training. Finally, we perform numerical experiments on Poisson and wave equations on irregular domains. We illustrate the generalizability, accuracy, and efficiency of the proposed algorithms on different PDE parameters, numbers of collection points, and several types of RBFs.

\end{abstract}


\end{frontmatter}


\section{Introduction}

Partial differential equations (PDEs), especially spatiotemporal PDEs, have been extensively used in several fields, such as physics, biology, and finance. However, except for some simple equations for which analytical solutions exist, solving PDEs is a challenging problem. Consequently, numerical approaches, including the finite element (FEM), finite volume (FVM), and finite difference (FDM), were developed to solve PDEs in various practical problems. 

In recent years, the rise of deep learning has provided an alternative solution for complex nonlinear PDEs without the need to use domain discretization in numerical methods. A pioneering work in this direction is the physics-informed neural networks (PINNs) \cite{raissi2019physics}, which constrain the output of deep neural networks to satisfy the PDEs via minimizing a loss function. PINNs have emerged as a promising framework for exploiting information from observational data and physical equations and can be classified into two categories: continuous and discrete. The continuous PINNs are to build a map from the domain to the solution by a feed-forward multi-layer neural network, where the partial derivatives can be easily computed through automatic differentiation (AD) \cite{2015AutomaticML}. It has been widely used in solving several engineering applications, such as fluid flow \cite{highspeedflows,2020Physics}, or solid mechanics \cite{2021Asolidmechanics}. While the continuous PINNs still have some limitations. First, a large number of points are required to represent the high-dimensional domain, and AD requires saving the differential computation map during training, which significantly increases the training cost and computation time. Second, the residual form of the PDE and its initial (IC) and boundary conditions (BC) are reduced to a composite objective function as an unconstrained optimization problem. This leads to the fact that it is difficult to enforce IC and BC for continuous PINNs strictly. What's more, the use of fully connected networks would limit the fitting accuracy of PINNs.

To raise the representation and effectiveness, the discrete PINNs that employ numerical discretizations to compute the derivative terms of physical information loss have attracted significant attention. Chen et al. \cite{2021HCP} discretized the computational domain by a regular mesh and used the FDM to discrete the PDE. Furthermore, proposed the theory-guided hard constraint projection (HCP) to define the PDE loss function. In addition to FD-based PINNs, CAN-PINN \cite{chiu2021canpinn} based on the Coupled-Automatic-Numerical Differentiation Method has been presented. To address parametric PDEs with unstructured grids, several recent works have been devoted to constructing generalized discrete loss functions based on FVM \cite{Rezaei2022}, or FEM \cite{Jot} and integrating them into physics-based neural network algorithms. In addition to numerical discretization, convolutional neural networks (CNNs) are often used in discrete PINNs. Zhu et al. \cite{ZHU201956} demonstrated that CNN-based discrete PINNs have higher computational efficiency when solving high-dimensional elliptic PDEs. Cai et al. \cite{cai8793167} investigated the CNN named LiteFlowNet to solve the fluid motion estimation problem. Further, Fang et al. \cite{fang9403414} developed hybrid PINNs based on CNNs and FVM to solve PDEs on arbitrary geometry. 

However, due to the inherent limitations of classical CNN convolution operations, it remains challenging for CNN-based discrete PINNs to handle irregular domains with unstructured grids. We can address the problem with graph neural networks (GNNs). Because graph convolution operates in non-Euclidean local space, it allows the network to learn the evolution of spatial localization, which is consistent with physical processes and has better interpretability. Jiang et al. \cite{Jiang} proposed the PhyGNNet that solves spatiotemporal PDEs with Physics-informed Graph Neural Networks, specifically, using FDM for physical knowledge embedding. Gao et al. \cite{GAO2022114502} presented a novel discrete PINN framework based on GNN and used the Galerkin method that is meshless to construct the PDE residual. In general, GNN-based discrete PINNs have more desirable fitting ability and generalization performance.

In this work, we propose a physics-informed framework (RBF-MGN) based on GNNs and radial basis function finite difference (RBF-FD) to solve spatio-temporal PDEs. The contributions are summarized as follows:

(a) We introduce graph convolutional neural networks into physics-informed learning to better handle irregular domains with unstructured meshes. We choose MeshGraphNets \cite{Mesh}, a graph neural network model with an Encoder-Processer-Decoder architecture, to model the discretized solution.

(b)  Radial basis function finite difference (RBF-FD), a meshless method, is used to process the model output node solution maps and construct a high-precision difference format of the differential equations to guide model training. Moreover, ensure that the output fully satisfies the underlying boundary conditions.

(c) We conduct several experiments on Poisson and wave equations, which indicates that our method has excellent ability and extrapolates well on irregular domains.

The rest of the paper is structured as follows. Section 2 provides a detailed introduction to the GNN, RBF-FD, and the principle of the RBF-MGN method. In Section 3, we provide o numerical results showcasing the performance of the proposed approach. In Section 4, we conclude this work and extensions to address the limitations.


\section{Methods}
\subsection{Overview}
Consider a dynamic physical process governed by general nonlinear and time-dependent PDEs of the form:
\begin{equation}
\begin{array}{c}
\boldsymbol{u}_{t}+\mathcal{L}[\boldsymbol{u},\eta]=0, \quad x \in \Omega, t \in[0, T], \\
\boldsymbol{u}(x, 0)=\boldsymbol{h}(x), \quad x \in \Omega, \\
\boldsymbol{u}(x, t)=\boldsymbol{g}(x, t), \quad x \in \partial \Omega, t \in[0, T],
\end{array}
\label{nonlinear PDE}
\end{equation}
where $\Omega \subset \mathbb{R}^{d} $ and $t \in[0, T]$ denote the computational domain and time coordinates. $\mathcal{L}$ represents the spatial-temporal differential operator. $\eta$ is the PDE parameter vector. The set of PDEs is subjected to the initial condition $\boldsymbol{h}(x)$ and boundary condition $\boldsymbol{g}(x,t)$, which is defined on the boundary $\partial \Omega$ of the domain. 

In this paper, we propose an innovative physics-informed graph neural network (RBF-MGN) to seek a solution function $\boldsymbol{u}(x,t)$ under the IC and BC. In the framework, we generate an unstructured grid and regard the mesh as a graph to train the GNN. The GNN aims to put the solution at time $t$ to obtain the solution at the next time step $t + \Delta t$. The IC is first used as an input to the model at the very beginning. The loss function is calculated based on the RBF-FD method associating the BC and the model output. The following subsections describe each component of the proposed method in detail.

\begin{figure}[htbp]
	\centering
	\subfigure{
		\includegraphics[scale=0.35]{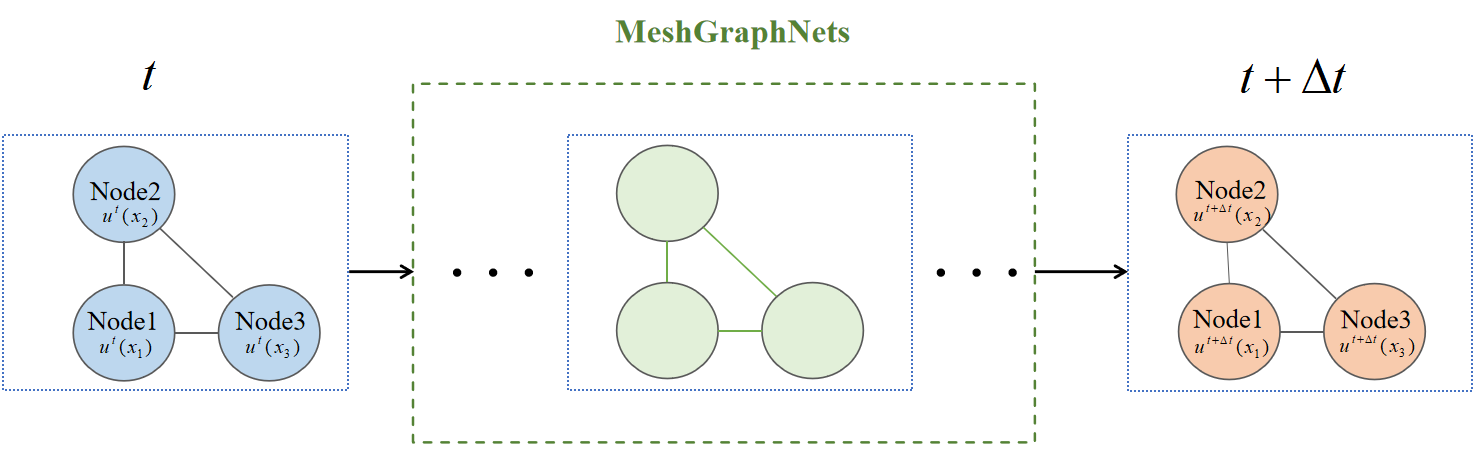}
	}
	\caption{An example of a GNN, given the input/output graph $G = (V, E)$, where V and E are a set of vertices $V = {1, 2, 3}$ and edges $E \subseteq\left(\begin{array}{l}V \\ 2\end{array}\right)$. And the same adjacency matrix $(N(1) = {2, 3}, N(2) = {1, 3}, N(3) = {1, 2})$. The input and out features are the nodal solution $i$ at time $t$ vector ( $f^{(in)}_i = u^{t+\delta t} (x_i ))$ )and at time $t+\Delta t$ vector ( $f^{(out)}_i = u^{t+\delta t} (x_i ))$.}
	\label{Graph}
\end{figure} 
\subsection{Graph neural networks}
As an emerging technology for flexible processing of unstructured data in deep learning, Graph neural networks (GNNs) have been widely used to solve various scientific machine learning problems.
A mesh with unstructured grids and corresponding nodal PDE solutions can be naturally described as graphs. The task is to represent the GNNs approximation of the solution of the equation \eqref{nonlinear PDE} at time $t+\Delta t$ given the current solution.

\subsubsection{Graphs}
First, we generate an irregular mesh and express the mesh as a graph $G = (V, E)$ with nodes $V$ connected by edges $E$. Each node $i \in V $ is defined by its feature vector, and the adjacent nodes are connected via edges. In the framework, the input graph of GNN is that each node is associated with its current PDE solution at time $t$, and then the output graph is the solution at time $t+\Delta t$ shown in Fig.\ref{Graph}.
\subsubsection{MeshGraphNets}
\begin{figure}[htbp]
	\centering
	\subfigure{
		\includegraphics[scale=0.45]{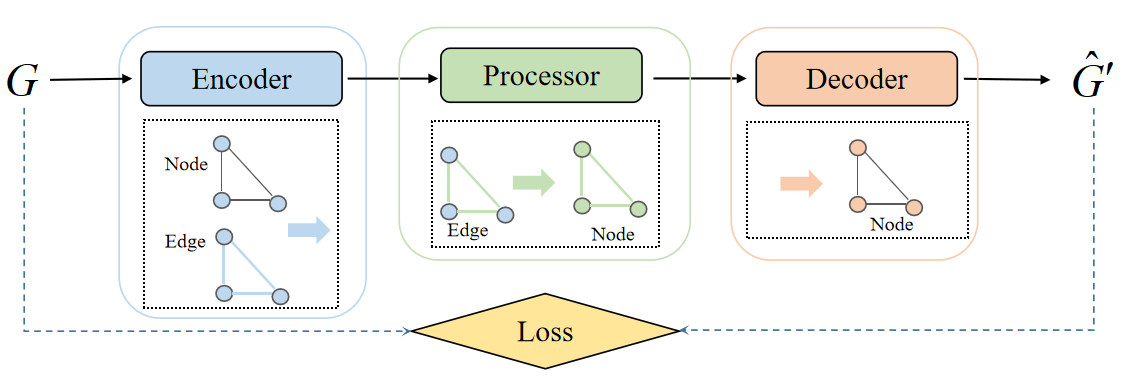}
	}
	\caption{Diagram of MeshGraphNets.}
	\label{MeshGraphNets}
\end{figure}

We use MeshGraphNets \cite{Self-Adaptive}, a graph neural network model with an Encoder-Processer-Decoder architecture, to model the discretized solution. The framework of the MeshGraphNets is constructed as shown in Fig.\ref{MeshGraphNets}, which mainly has three parts.

Firstly, the Encoder encodes features into graph nodes and edges. And the Encoder has two hidden layers with the ReLU activation function, and each layer has 128 hidden units. Secondly, The processer predicts latent feature variation of nodes via a Graph Network(GN) that updates a graph state, including the attributes of the node, edge, and whole graph. Each block contains a separate set of network parameters and is applied in sequence to the output of the previous block, updating the edge  $e_{ij}$ and then node  $v_{i}$. Finally, the decoder decodes node features with the MLP of the same architecture as the Encoder as correction of the input to create final predicts. When training, the losses are computed to update the network parameters $\boldsymbol{\Theta}$ at once.

\begin{figure}[htbp]
	\centering
	\subfigure[FD]{
		\includegraphics[scale=0.45]{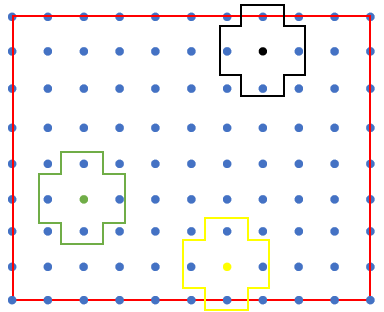}
	}%
	\subfigure[RBF-FD]{
		\includegraphics[scale=0.45]{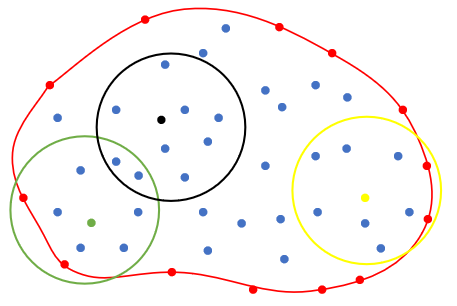}
	}%
	\centering
	\caption{The corresponding local region of (a) The classical difference. (b) The RBF-FD method.}
	\label{RBF-FD}
\end{figure}
\subsection{PDE-informed loss function}
The GNN training requires the known differential equations to be enforced by the loss function built on the basis of PDE residuals. The above PDE consists of several derivative operators, such as  $\boldsymbol{u}_{t}$, $\nabla u$, and $\Delta u$. In continuous PINN, the spatial and temporal gradient operators of PDEs are computed using the available technology of AD. However, the method is essentially a soft constraint whose regularization term in the loss function can only guarantee that the predicted results do not severely violate the constraint in an average sense and may still produce physically inconsistent results. A hard constraint approach must be proposed to ensure that the PDE is strictly satisfied in the computational domain. In this work, a hard constraint approach based on the radial basis function finite difference (RBF-FD) technique is considered to embed domain knowledge into neural networks.

\begin{table*}[!t]
	\begin{center}
		\caption{\label{tab1} Common Radial Basis Functions.}
		\begin{tabular}{cccccccccccc} \toprule
			RBF  & $\phi(r)$   \\ \hline
			Gaussian (GA)  & $exp^{-(\varepsilon r)^2}$  \\
			inverse multiquadric (IMQ)  & $\frac{1}{\sqrt{1+(\varepsilon r)^2}}$  \\
			3rd order polyharmonic spline (ph3)  & $(\varepsilon r)^3$\\

			\bottomrule
		\end{tabular}
	\end{center}
\end{table*}

\subsubsection{RBF-FD}
Tolsrykh et al. \cite{Tolstykh} first discussed the radial basis function difference method (RBF-FD) that applies radial basis functions to finite differences. The RBF-FD method belongs to the meshless method, which makes it easy to handle problems in irregular areas and scattered node layouts. It needn’t mesh generation, and without numerical integration that could save a lot of computing time. RBF-FD method has been applied to numerical solutions in many scientific and engineering fields, for example, incompressible flow and heat conduction problems \cite{RBF-FD}.

The RBF-FD method is a method for spatial discretization of differential operators on spatial scatter. Assuming there are $n$ collection points $\left\{\mathbf{x}_i\right\}_{i=1}^n$ on the domain $\Omega$, and we choose the $m$ nearest neighbor points to form the corresponding local region $\left\{\mathbf{x}_i^k\right\}_{k=1}^m=\Omega_i$ for each point $\mathbf{x}_i$. The classical difference method approximates the solution $u(\mathbf{x})$ as a polynomial function on a regular local grid shown in Fig.\ref{RBF-FD}(a) and represents the differentiation of the solution as a weighted sum of the function values of several grid nodes by Taylor expansion. The RBF-FD method approximates the solution $u(\mathbf{x_i})$ in local irregular space $\Omega_i$ shown in Fig.\ref{RBF-FD}(b) as a combination of radial basis functions $\phi\left(x_i, x_{i}^{k}\right)$ and polynomial functions $p(x_i)$. 
\begin{equation}
u(x_i) \approx \sum_{k=1}^{m} \lambda_{k} \phi\left(x, x_{i}^{k}\right)+\sum_{k=1}^{q} \mu_{k} p_{k}(x_i),
\label{ux}
\end{equation}
where $\lambda_{k}$ and $\mu_{k}$ are the corresponding combination coefficients. And $q$ is the number of terms of the polynomial, with the constraint conditions,
$\sum_{i=1}^{n} \lambda_{i} p_{j}(x_i)=0, j=1,...,q$. Combining the above equations, we could obtain,
\begin{equation}
\begin{gathered}
\left[\begin{array}{cc}
\boldsymbol{A} & \boldsymbol{P} \\
\boldsymbol{P}^{\mathrm{T}} & \mathbf{0}
\end{array}\right]\left[\begin{array}{l}
\lambda \\
\mu
\end{array}\right]=\left[\begin{array}{l}
\boldsymbol{u} \\
\mathbf{0}
\end{array}\right]\\
\boldsymbol{A}=\left[\begin{array}{cccc}
\phi\left(x_{1}, x_{1}\right) & \phi\left(x_{2}, x_{1}\right) & \cdots & \phi\left(x_{m}, x_{1}\right) \\
\phi\left(x_{1}, x_{2}\right) & \phi\left(x_{2}, x_{2}\right) & \cdots & \phi\left(x_{m}, x_{2}\right) \\
\vdots & \vdots & \ddots & \vdots \\
\phi\left(x_{1}, x_{m}\right) & \phi\left(x_{2}, x_{m}\right) & \cdots & \phi\left(x_{m}, x_{m}\right)
\end{array}\right] \\
\boldsymbol{P}=\left[\begin{array}{cccc}
p_{1}\left(x_{1}\right) & p_{2}\left(x_{1}\right) & \cdots & p_{q}\left(x_{1}\right) \\
p_{1}\left(x_{2}\right) & p_{2}\left(x_{2}\right) & \cdots & p_{q}\left(x_{2}\right) \\
\vdots & \vdots & \ddots & \vdots \\
p_{1}\left(x_{m}\right) & p_{2}\left(x_{m}\right) & \cdots & p_{q}\left(x_{m}\right)
\end{array}\right] \\
\mu=\left[\begin{array}{llll}
\mu_{1} & \mu_{2} & \cdots & \mu_{q}
\end{array}\right]^{\mathrm{T}} \\
\lambda=\left[\begin{array}{llll}
\lambda_{1} & \lambda_{2} & \cdots & \lambda_{m}
\end{array}\right]^{\mathrm{T}} \\
\boldsymbol{u}=\left[\begin{array}{llll}
u_{1} & u_{2} & \cdots & u_{m}
\end{array}\right]^{\mathrm{T}}
\end{gathered}
\end{equation}
Therefore, the corresponding combination coefficients are as follows:
\begin{equation}
\left[\begin{array}{l}
\lambda \\
\mu
\end{array}\right]=\left[\begin{array}{cc}
\boldsymbol{A} & \boldsymbol{P} \\
\boldsymbol{P}^{\mathrm{T}} & \mathbf{0}
\end{array}\right]^{-1}\left[\begin{array}{l}
\boldsymbol{u} \\
\mathbf{0}
\end{array}\right].
\label{the corresponding combination coefficients}
\end{equation}
Further, consider the differential operator $\mathcal{L}$ on the solution $u(x)$, $\left.\mathcal{L} u(x)\right|_{x=x_{i}}$ at the space point $x_i$ can be approximated as: 
\begin{equation}
\left.\mathcal{L} u(x)\right|_{x=x_{i}} \approx \sum_{k=1}^{m} w^{i}_{k} u_{x_{i}^{k}},
\label{operator}
\end{equation}
where $w^{i}_{k}$ are differentiation weights for the point $x_i$. To determine $w^{i}_{k}$, Substitute the equation \eqref{ux} into the above equation:
\begin{equation}
\sum_{k=1}^{m} \lambda_{k} \mathcal{L} \phi\left(x, x_{i}^{k}\right)+\sum_{k=1}^{q} \mu_{k} \mathcal{L} p_{k}(x_i) \approx \sum_{k=1}^{m} w^{i}_{k} u_{x_{i}^{k}},
\end{equation}
Performing matrix decomposition yields,
\begin{equation}
\begin{aligned}
&\left[\begin{array}{ll}
\boldsymbol{b} & \boldsymbol{c}
\end{array}\right]\left[\begin{array}{l}
\lambda \\
\mu
\end{array}\right]  =\left[\begin{array}{ll}
w & v
\end{array}\right]\left[\begin{array}{l}
u \\
0
\end{array}\right]\\
\boldsymbol{b} &=\left[\begin{array}{llll}
\mathcal{L} \phi\left(x_{i}, x_{1}\right) & L \phi\left(x_{i}, x_{2}\right) & \cdots & L \phi\left(x_{i}, x_{m}\right)
\end{array}\right] \\
\boldsymbol{c} &=\left[\begin{array}{llll}
\mathcal{L} p_{1}\left(x_{i}\right) & L p_{2}\left(x_{i}\right) & \cdots & L p_{m}\left(x_{i}\right)
\end{array}\right] \\
\boldsymbol{w} &=\left[\begin{array}{llll}
w_{1} & w_{2} & \cdots & w_{m}
\end{array}\right] \\
\boldsymbol{v} &=\left[\begin{array}{llll}
v_{1} & v_{2} & \cdots & v_{q}
\end{array}\right]
\end{aligned}
\end{equation}
Substitute Eq. \eqref{the corresponding combination coefficients} into the above equation, we obtain all $w^{i}_{k}$:
\begin{equation}
\left[\begin{array}{l}
\boldsymbol{w}^{\mathrm{T}} \\
\boldsymbol{v}^{\mathrm{T}}
\end{array}\right]=\left[\begin{array}{cc}
\boldsymbol{A} & \boldsymbol{P} \\
\boldsymbol{P}^{\mathrm{T}} & \mathbf{0}
\end{array}\right]^{-1}\left[\begin{array}{l}
\boldsymbol{b}^{\mathrm{T}} \\
\boldsymbol{c}^{\mathrm{T}}
\end{array}\right].
\end{equation}
and substituting into Eq. \eqref{operator} get the approximate solution of $\mathcal{L}u(x)$ at the point $x_i$. Further, the approximate solution of $\mathcal{L}u(x)$ can be obtained for each point in the irregular computational domain.

\subsubsection{PDE residuals}
In this work, given $u^l(x)=u(x, t^{l})$ and the network output $u^{l+1}(x)=u(x, t^{l+1})$, we construct loss function with the RBF-FD method. The equation \eqref{nonlinear PDE} can represent a wide range of time-dependent PDEs, such as the Poisson equation and wave equation. 
Here, we give a heat transfer example $u_t = \alpha \Delta u$ to demonstrate how to define PDE residuals using RBF-FD. First, we sample $n_c$ collocation points on the domain $\Omega$ and $n_b$ boundary nodes, where $n = n_c + n_b$. And denote $\tau = t^{l+1}-t^{l}$ be the time step. 

For operator $u_t$, it can be approximated with backward difference on time $t$:
\begin{equation}
u_t = \frac{u^{l+1}(x)-u^{l}(x)}{\tau}
\end{equation}
According to equation \eqref{operator}, the Laplace item $\Delta u$ at the space point $x_i$ can be approximated as:
\begin{equation}
\Delta u^l(x) \approx \sum_{k=1}^{m} w^{i}_{k} u(x_{i}^{k})
\label{Delta}
\end{equation}
Therefore, the heat equation is discretized as the following formulation,
\begin{equation}
\alpha \sum_{k=1}^m \omega_k^i u^l\left(\mathbf{x}_i^k\right)+\frac{1}{\tau} u^l\left(\mathbf{x}_i\right)=\frac{1}{\tau} u^{l+1}\left(\mathbf{x}_i\right), \quad i=1,2, \ldots, n_c
\end{equation}
We consider the essential boundary conditions $u^l(\mathbf{x}_i)=h(\mathbf{x}_i, t^l), \quad i=n_c +1, n_c+2, \ldots, n$ and the discretized PDE. Transform the problem into the following linear algebraic equations:
\begin{equation}
\begin{aligned}
&\mathrm{A} U^l=\frac{1}{\tau} U^{l+1}+H^l\\
&a_{i j}= \begin{cases}\alpha \omega_{k}^{i} & i \neq j, \text { for } i=1,2, \ldots, n_{c}, \\ \alpha \omega_{k}^{i}+\frac{1}{\tau} & i=j, \text { for } i=1,2, \ldots, n_{c}, \\ 1 & i=j, \text { for } i=n_{c}+1, n_{c}+2, \ldots, n\end{cases}\\
&U^{l}=\left[\begin{array}{lllll}
u^{l}\left(\mathbf{x}_{1}\right) & u^{l}\left(\mathbf{x}_{2}\right) & \ldots & u^{l}\left(\mathbf{x}_{n}\right)
\end{array}\right]^{T}, \\
&H^{l}=\left[\begin{array}{lllllll}
0 & \ldots & 0 & h^{l}\left(\mathbf{x}_{n_{c}+1}\right) & h^{l}\left(\mathbf{x}_{n_{c}+2}\right) & \ldots & h^{l}\left(\mathbf{x}_{n}\right)
\end{array}\right]^{T},\\
&U^{l+1}=\left[\begin{array}{lllll}
u^{l+1}\left(\mathbf{x}_{1}\right) & u^{l+1}\left(\mathbf{x}_{2}\right) & \ldots & u^{l+1}\left(\mathbf{x}_{n_c}\right)
\end{array}\right]^{T}, 
\end{aligned}
\label{AMatrix}
\end{equation}
where $U^{l}$ and $\hat{U}^{l+1}$ are the prediction matrix. $\mathrm{A}$ is the constraint matrix that denotes the physical constraints. $H^l$ is the boundary matrix.
The solution $\hat{U}^{l+1}$ will be learned by GNN as the output graph $\hat{U}^{l+1}(\boldsymbol{\Theta})$. The PDE residual is as follows:
\begin{equation}
\boldsymbol{R}_u\left(\hat{U}^{l+1}(\boldsymbol{\Theta}), U^l ; \boldsymbol{\alpha}\right) = \mathrm{A} U^l-\frac{1}{\tau} \hat{U}^{l+1}-H^l,
\label{PDE residual}
\end{equation}
The PDE-informed loss function for the GNN has the following form, the essential boundary condition will be satisfied automatically:
\begin{equation}
\begin{aligned}
\mathcal{L}_{\mathrm{f}}(\boldsymbol{\Theta})=\left\|\boldsymbol{R}_u\left(\hat{U}^{l+1}(\boldsymbol{\Theta}), U^l ; \boldsymbol{\alpha}\right)\right\|_2 .
\label{loss}
\end{aligned}
\end{equation}
Then we can use Adam to minimize the loss $\mathcal{L}_{\mathrm{f}}(\boldsymbol{\Theta})$ as close to zero as possible to adjust the network weight $\boldsymbol{\Theta})$. Solving heat transfer problem with RBF-MGN is summarized as the algorithm\ref{alg:RBF-MGN}.

\begin{algorithm}[htb]
	\caption{Solving heat transfer problem with RBF-MGN}
	\label{alg:RBF-MGN}
	
	\begin{algorithmic}
		\State \textbf{Step\hspace{0.5em}1}:Generate an unstructured grid and regard the mesh as a graph $G = (V, E)$.
		\State \textbf{Step\hspace{0.5em}2}:Construct MeshGraphNets to put the matrix $U^{l}$ and  at time $t$ to obtain prediction matrix $\hat{U}^{l+1}$ at next time step $t + \Delta t$. The IC is first used as an input to the model at the very beginning.
		\State \textbf{Step\hspace{0.5em}3}:Compute the constraint matrix $\mathrm{A}$ (Eq. \eqref{AMatrix}) based on the RBF-FD. Obtain the boundary matrix $H^l$ with BC.
		\State \textbf{Step\hspace{0.5em}4}:Formulate the the PDE residual (Eq.\eqref{PDE residual}). 
		\State \textbf{Step\hspace{0.5em}5}:Solve the optimization problem (Eq.\eqref{loss}) to obtain the next state solution.
	\end{algorithmic}
\end{algorithm}

\section{Results}
In this section, we will present several numerical experiments to show the ability of the proposed method for solving the PDEs especially on complex domain, include solving the two-dimensional Poisson’s equation and two-dimensional wave equation. We also consider different time steps $\tau$, PDE parameters and several types of RBFs on the learning performance of the proposed method. We further study the performance of RBF-MGN with different numbers of collection points $n$ and nearest neighbor nodes $m$. All numerical experiments are mainly based on Pytorch. The MLPs with two hidden layers, each with 64 neurons in the encoder, processor, and decoder of the neural network are employed in all experiments. The activation function is ReLU. Unless otherwise specified, The optimizer is Adam and the learning rate is set to 0.001. The rest of the detailed configurations are described in the respective experiment. In order to test the accuracy, two errors –the absolute error and the relative L2 error are used, which are defined as follows:
\begin{equation}
\begin{aligned}
\text {absolute error }=\left|\hat {u}-u\right|,\\
\text {relative L2 error }=\frac{\sqrt{\left|\hat {u}-u\right|^{2}}}{\sqrt{\left|u\right|^{2}}}.
\label{L2error}
\end{aligned}
\end{equation}

\subsection{Two-dimensional Poisson’s equation}
The first experiment considers a simple two-dimensional Poisson’s equation:
\begin{equation}
\begin{aligned}
u_{t}+\gamma \Delta u+f(x, y, t)=0,  \quad x \in [0, 1], y \in[0, 1].
\end{aligned}
\end{equation}

\begin{figure}[htbp]
	\centering
	\subfigure{
		\includegraphics[scale=0.25]{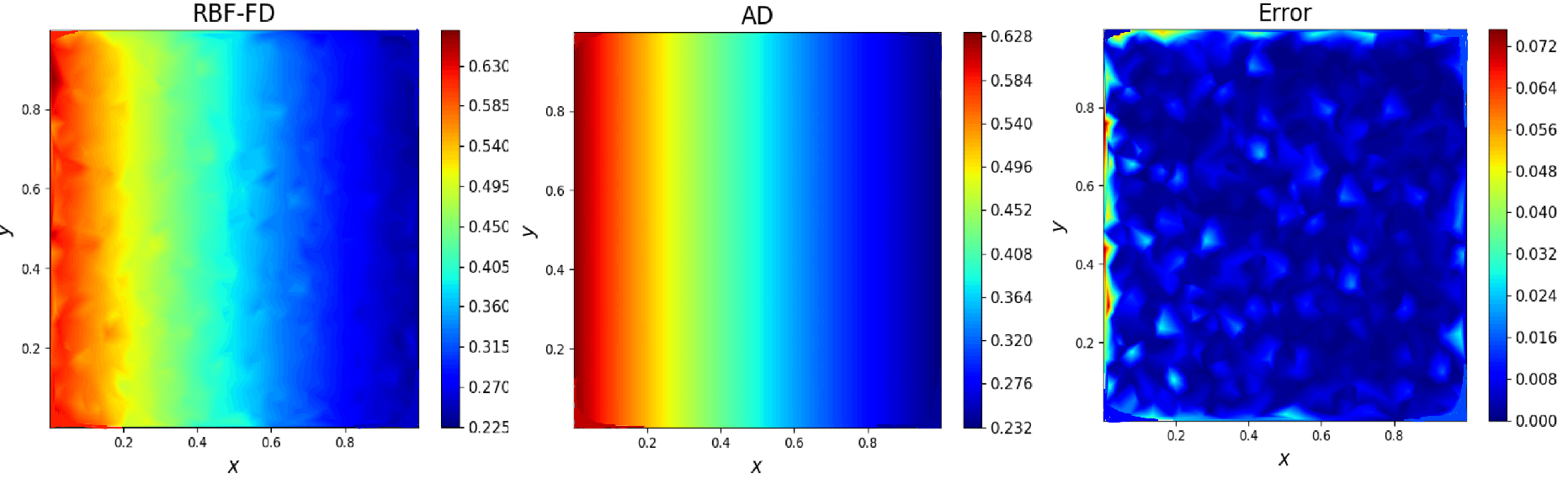}
	}
	\caption{The Laplace operator $\Delta u$ approximated using RBF-FD \eqref{Delta} and AD.}
	\label{Square-delta}
\end{figure}

First, assuming that $f(x, y, t)$ is 0, the Laplace operator $\Delta u$ approximated using Eq. \eqref{Delta} and AD are shown in Fig.\ref{Square-delta}. In the computation, we choose unstructured $n=167$ collection points, $m = 10$ nearest neighbor nodes. We use the ph3 RBF with shape parameter $\varepsilon=1$ to compute the weights, and the order of the added polynomial is 2. It can be easily seen that RBF-FD is a perfect substitute for AD to approximate the Laplace operator, which in turn can be used to define the PDE residuals.

\begin{figure}[htbp]
	\centering
	\subfigure{
		\includegraphics[scale=0.25]{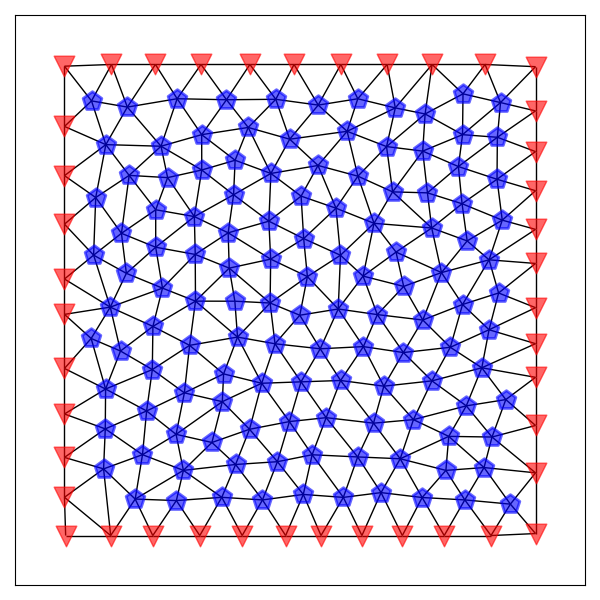}
	}
	\caption{The graph $G = (V, E)$ with nodes $V$ connected by edges $E$, V is a set of 167 points on a two-dimensional domain , including boundary nodes (red triangles), interior nodes (blue pentagons).}
	\label{Square}
\end{figure}

Assume that the boundary conditions are Dirichlet and $f(x, y, t)=-3-2\gamma(x+y)$, we take $\gamma=1$ to obtain the analytical solution $u(x, y, t)=xy^2+yx^2+3t$. We represent the computational domain as a graph $G=(V, E)$ with the simple structured Delaunay triangulation in 2D with the Bowyer-Watson algorithm shown in Fig.\ref{Square}. V is a set of points on a two-dimensional domain, including boundary nodes (red triangles) and interior nodes (blue pentagons). The edge e is a closed line segment formed by the points in the set of points as endpoints, and E is the set of e. 

\begin{figure}[htbp]
	\centering
	\subfigure[PDE residual]{
		\includegraphics[scale=0.25]{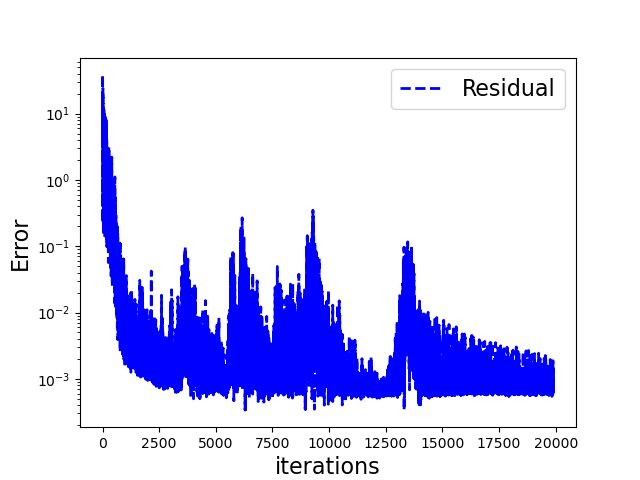}
	}%
	\subfigure[Train error]{
		\includegraphics[scale=0.25]{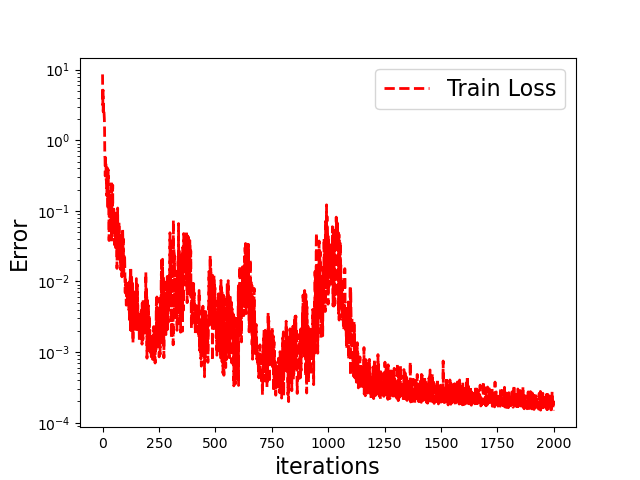}
	}%
	\centering
	\caption{Two-dimensional Poisson’s equation: The results of the PDE residual (a) and train error (b) on different spatial areas along with training iterations.}
	\label{Square-loss}
\end{figure}

\begin{figure}[htbp]
	\centering
	\subfigure{
		\includegraphics[scale=0.1]{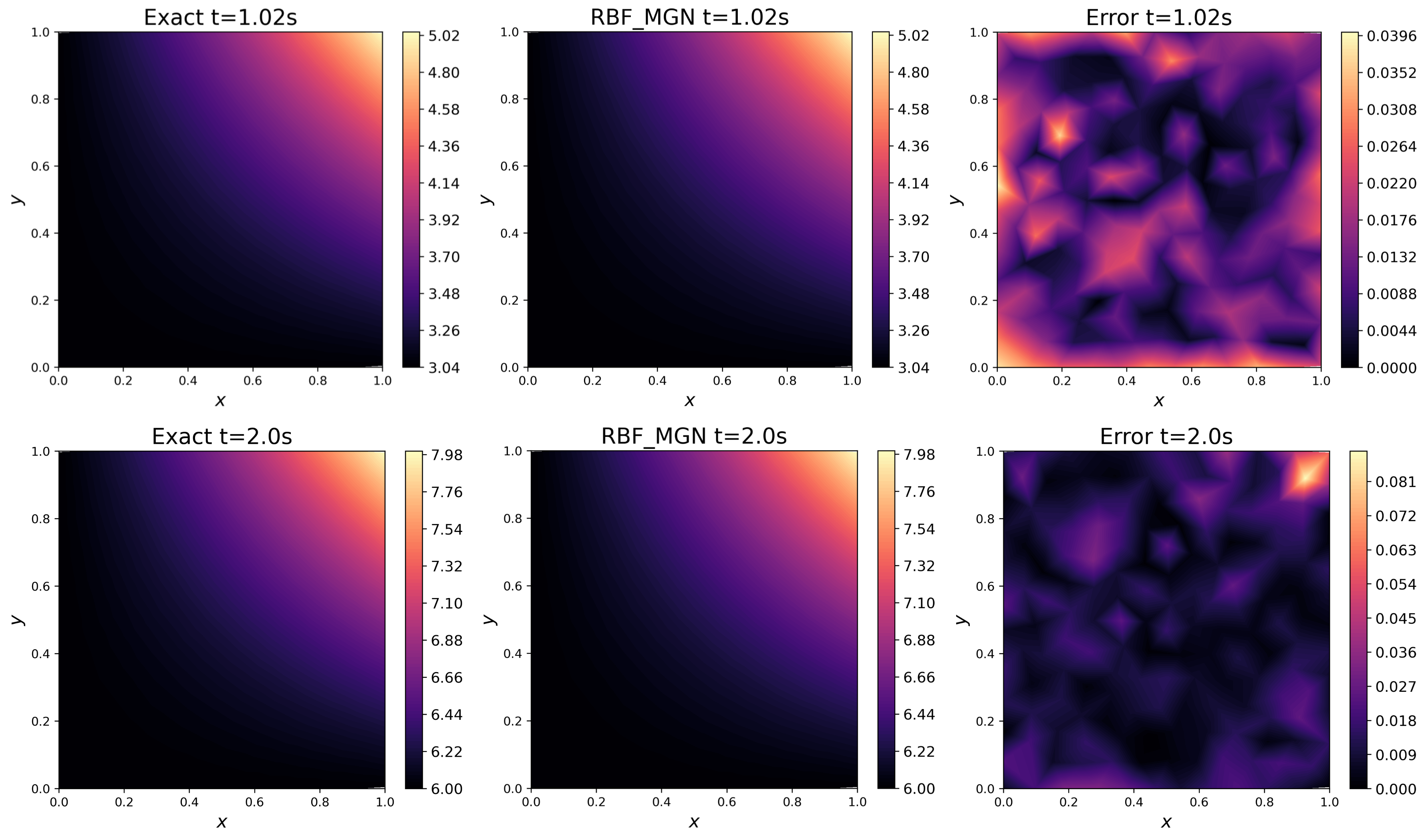}
	}
	\caption{The reuslts of two-dimensional Poisson’s equation at different time steps. The predicted results are compared with the exact solutions and the difference is also presented.}
	\label{Square-error}
\end{figure}

First, we consider the IC as the exact solution at time $t=0$ and train the network to infer the solution at $t \in[0, T], T=1$ with the time step $\tau = 0.01s$. For the RBF-FD method, we use the ph3 RBF, and Analogous to the Eq. \eqref{PDE residual}, the PDE residual is $\mathrm{A} U^l-\frac{1}{\tau} \hat{U}^{l+1}-H^l + F^l$ where $F^l$ is the matrix associated with $f(x, y, t)$. As shown in Fig.\ref{Square-loss}, the residual based on the RBF-FD definition converges quickly, guaranteeing that the prediction relative errors are all less than 0.001. The predicted and exact solutions at $t = 1.02, 2.0$ and the absolute errors are compared as shown in Fig.\ref{Square-error}. 

\begin{figure}[htbp]
	\centering
	\subfigure{
		\includegraphics[scale=0.1]{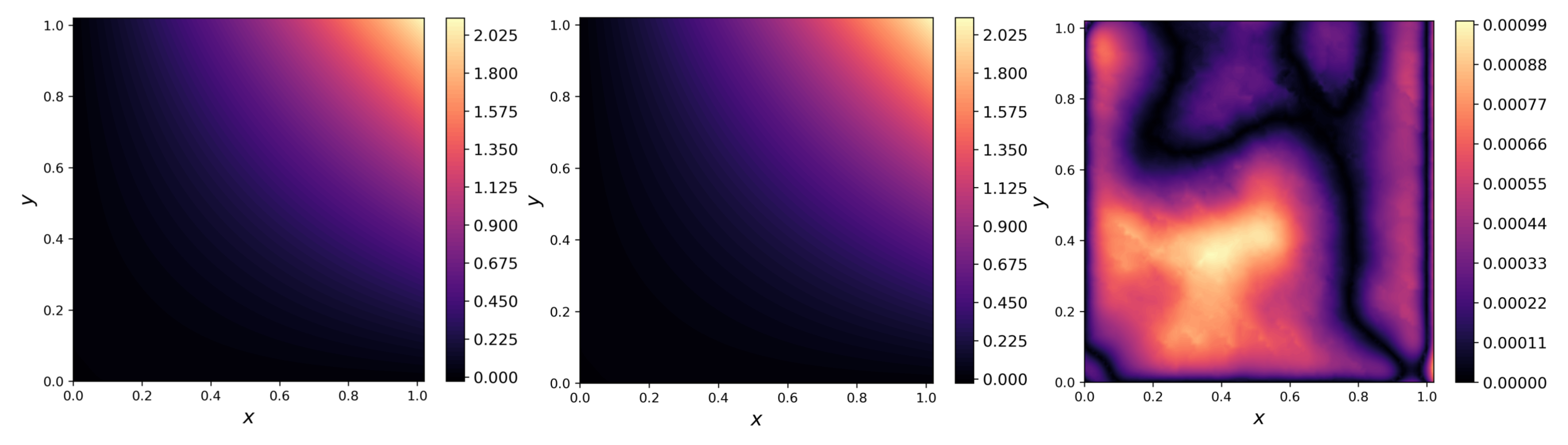}
	}
	\subfigure[Train error]{
		\includegraphics[scale=0.40]{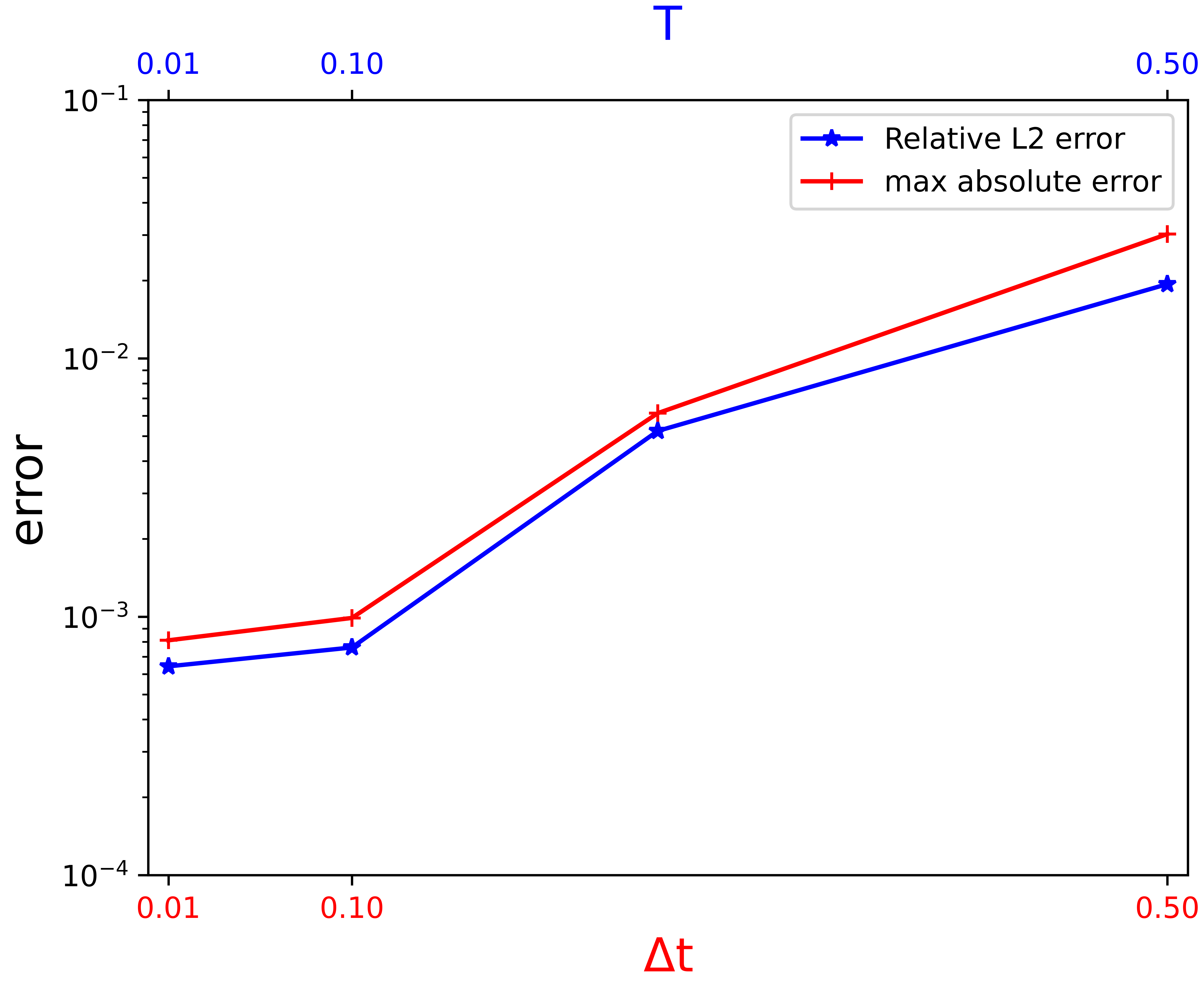}
	}%

	\caption{Two-dimensional Poisson’s equation:The graphs of the initial temperature reconstructed by RBF-MGN}
	\label{Square-inverse}
\end{figure}

In addition, we need to find the initial temperature from the final temperature at $T$. We employ the IMQ RBF to define the loss function. The graphs of exact and predicted solutions at the initial moment are presented in Fig.\ref{Square-inverse} with $\tau = 0.1$, and the two figures are almost identical. Furthermore, RBF-MGN could achieve excellent accuracy with different $tau$ as shown in Table \ref{tab2} and Fig.\ref{Square-inverse}.

\begin{table*}[!t]
	\begin{center}
		\caption{\label{tab2}Two-dimensional Poisson’s equation: The results of RBF-MGN with different $\tau$.}
		\begin{tabular}{cccccccccccc} \toprule
			$\tau$  &  the final T  & max absolute error  &  Relative L2 error \\ \hline
			{0.5}  & {0.5}  & {3.03e-02}  & {1.94e-02}\\
			{0.25}  & {0.25}  & {6.14e-03}  & {5.24e-03}\\
			{0.1}  & {0.1}  & {9.90e-04}  & {7.61e-04}\\
			{0.01}  & {0.01} & {8.11e-04}  & {6.43e-04}\\
			
			\bottomrule
		\end{tabular}
	\end{center}
\end{table*}

\subsection{Two-dimensional Poisson’s equation on amoeba domain}

In this example, we consider the heat transfer problem with the amoeba domain as follows:
\begin{equation}
\begin{aligned}
u_t &= \lambda \Delta u, \quad(x, y) \in \Omega,\\
\partial \Omega&=\{(x, y) \mid x=\rho \cos \theta+1, \quad y=\rho \sin \theta+1, \quad \theta \in[0,2 \pi]\},
\end{aligned}
\end{equation}
where $\rho=\left(\exp (\sin \theta) \sin ^2 2 \theta+\exp (\cos \theta) \cos ^2 2 \theta\right) / 2$. We use the analytical solution of the following form $u(x,y,t)=\lambda exp(-t)(cosx+cosy)$, and the initial condition is given as $u=\lambda (cosx+cosy)$. 

\begin{figure}[htbp]
	\centering
	\subfigure{
		\includegraphics[scale=0.15]{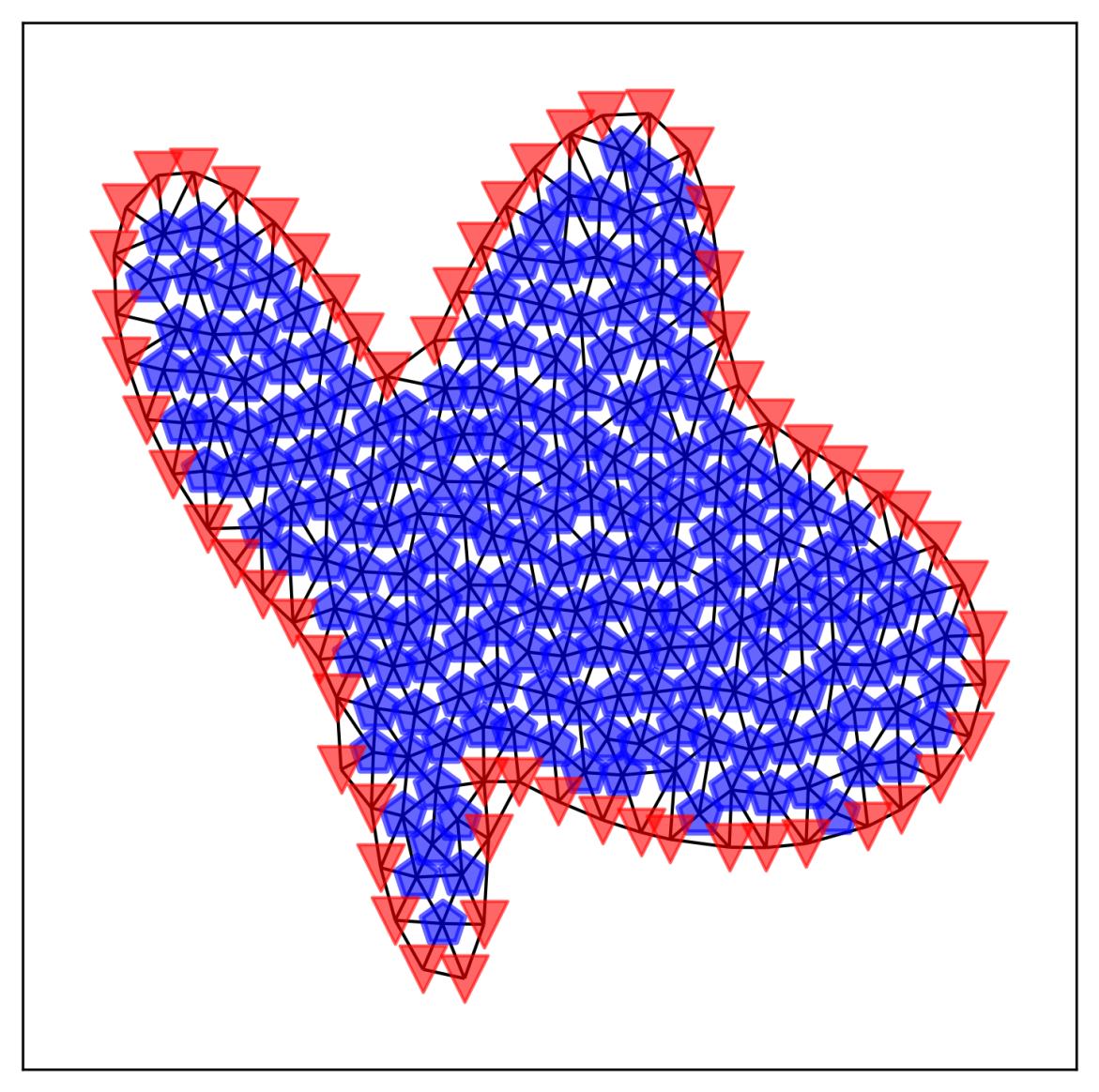}
	}
	\caption{The graph $G = (V, E)$ with nodes $V$ connected by edges $E$, V is a set of 167 points on a two-dimensional domain , including $n_b = 64$ boundary nodes (red triangles), $n_c = 195$ interior nodes (blue pentagons).}
	\label{amoeba}
\end{figure}

Triangulate the irregular region as shown in the Fig.\ref{amoeba} to represent the graph $G=(V, E)$, which includes $n_c = 195$ collocation points on and $n_b = 64$ boundary nodes. It is much lower than the total number of collocation points for a typical point-to-point PINN. We aim to attain the solution at $t \in[0, T], T=2$ with the time step $\tau = 0.01s$ through the RBF-MGN method. We construct data sets with a time range of 0 to 1 for training and test the ability of the model to infer solutions at $t \in[1, 2]$. For the RBF-FD method, we use ph3 RBF and sample $m = 15$ nearest neighbor nodes. 

\begin{figure}[htbp]
	\centering
	\subfigure[PDE residual]{
		\includegraphics[scale=0.25]{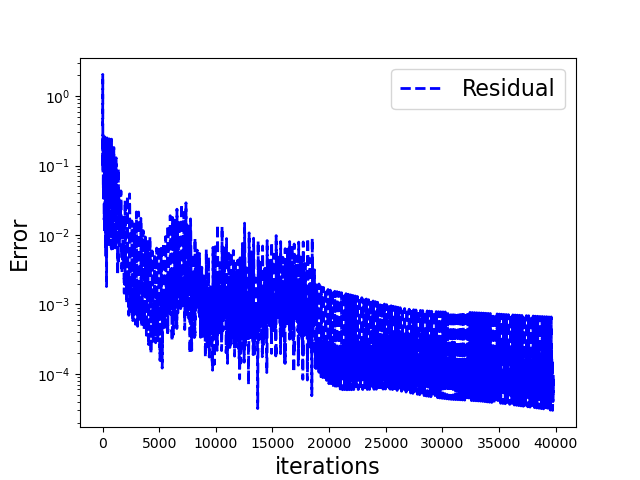}
	}%
	\subfigure[Train error]{
		\includegraphics[scale=0.25]{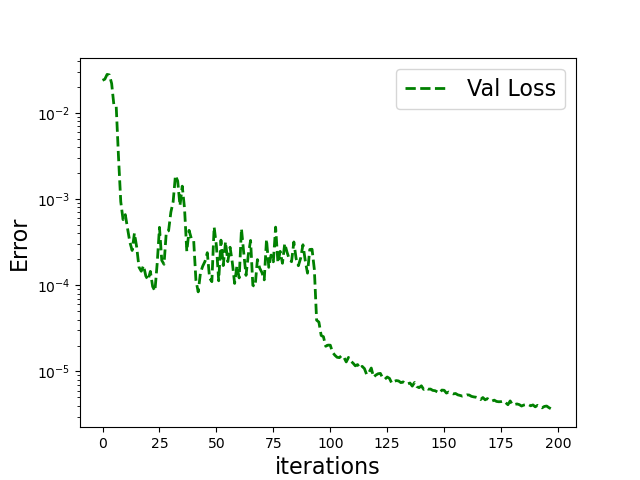}
	}%
	\centering
	\caption{Two-dimensional Poisson’s equation: The results of the PDE residual (a) and test error (b) on different spatial areas along with training iterations.}
	\label{amoeba-loss}
\end{figure}

\begin{figure}[htbp]
	\centering
	\subfigure{
		\includegraphics[scale=0.1]{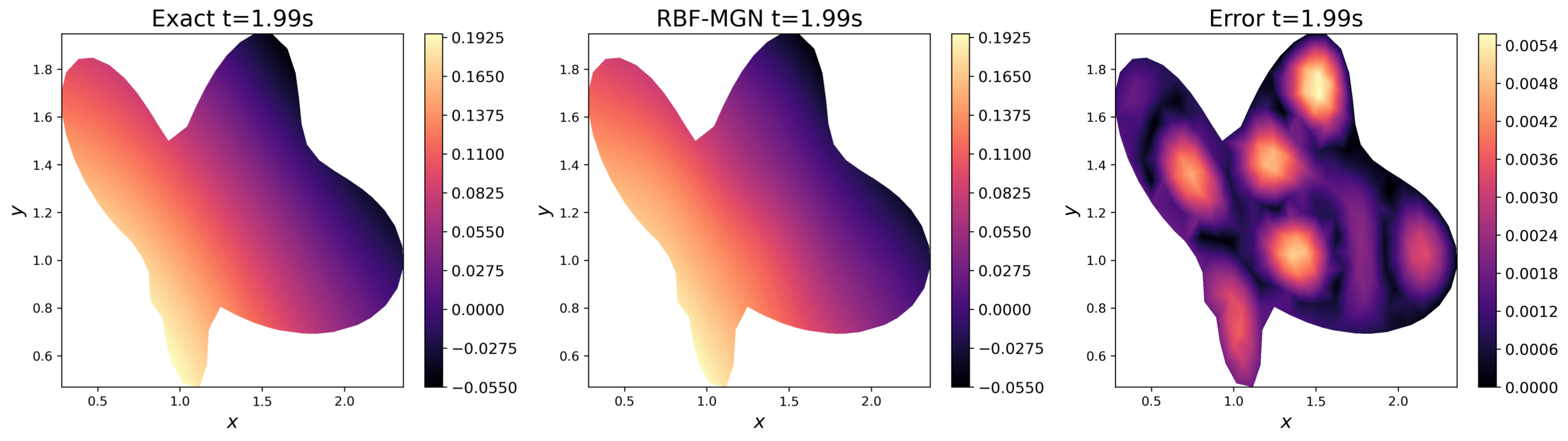}
	}
	\caption{The reuslts of two-dimensional Poisson’s equation on amoeba domain at $t = 1.99$. The predicted results are compared with the exact solutions and the difference is also presented.}
	\label{amoeba-error}
\end{figure}

Setting $\lambda = 1.0$, the batchsize is 5, and the fixed number of iterations is 200 Adam steps. As shown in Fig.\ref{amoeba-loss}, the residuals defined according to Eq.\eqref{PDE residual} are easy to handle, ensuring that the model predictions strictly conform to physical constraints. And then the val loss reaches to $1e-5$. RBF-MGN could accurately recover the temperature at $t = 1.99$, as shown in Fig.\ref{amoeba-error}. 

\begin{figure}[htbp]
	\centering
	\subfigure{
		\includegraphics[scale=0.40]{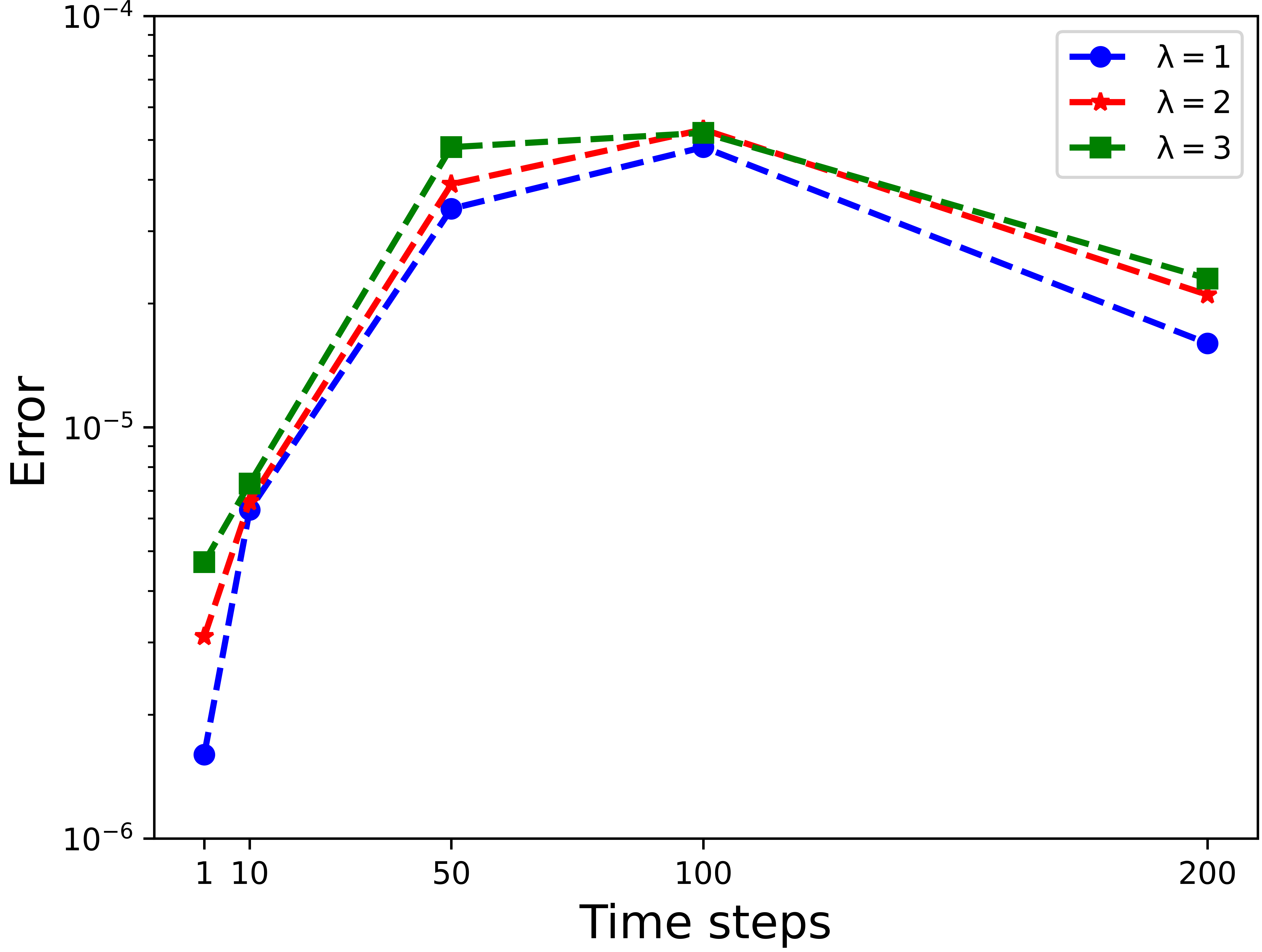}
	}
	\caption{The errors at different time steps of the heat transfer problem with the amoeba domain with different $\lambda$.}
	\label{amoeba-weight}
\end{figure}

In addition, the errors at different time steps of the heat transfer problem with the amoeba domain with different $\lambda$ are shown in Table \ref{tab3}, where the time step is fixed to $\tau = 0.01$. As we can see in Fig.\ref{amoeba-weight}, the error in step [1, 10] is on the 0.000001 level, which indicates that our approach could fit well even when extrapolating at time.

\begin{table} \fontsize{8}{8}\centering \caption{The errors at different time steps of the heat transfer problem on the amoeba domain with different $\lambda$.}\begin{tabular}{ | c | c | c | c | c | c | }\hline \diagbox{$\lambda$}{$setps$} & 1 & 10 & 50 & 100 & 200 \\\hline\hline 1 & {1.6e-6} & {6.3e-6} & {3.4e-5} &{4.8e-5} &{1.6e-5} \\\hline\hline 2 & {3.1e-6} & {6.6e-6} & {9.4e-6} &{5.3e-5} &{2.1e-5} \\\hline\hline 3 & {4.7e-6} & {7.3e-6} & {4.8e-5} &{5.2e-5} & {2.3e-5}\\\hline\end{tabular}\vspace{0cm}\label{tab3}\end{table}

\subsection{Two-dimensional Poisson’s equation on butterfly domain}
In the experiment, we also consider the problem in Example 2, but the analytical solution of the heat conduction equation is:
\begin{equation}
\begin{aligned}
u(x, y, t)=\exp \left(-\frac{\pi^2 t}{4}\right)\left[y \sin \left(\frac{\pi x}{2}-\frac{\pi}{4}\right)+x \sin \left(\frac{\pi y}{2}-\frac{\pi}{4}\right)\right].
\end{aligned}
\end{equation}

The initial and boundary conditions are obtained from the analytical solution. For the computation domain, we would use the butterfly is as follows:
$\Omega=\{(x, y) \mid x=0.55 \rho(\theta) \cos (\theta), y=0.75 \rho(\theta) \sin (\theta)\} \text { and } \rho(\theta)=1+\cos (\theta) \sin (4 \theta),0\leqslant \theta\leqslant 2\pi$. 

\begin{figure}[htbp]
	\centering
	\subfigure{
		\includegraphics[scale=0.15]{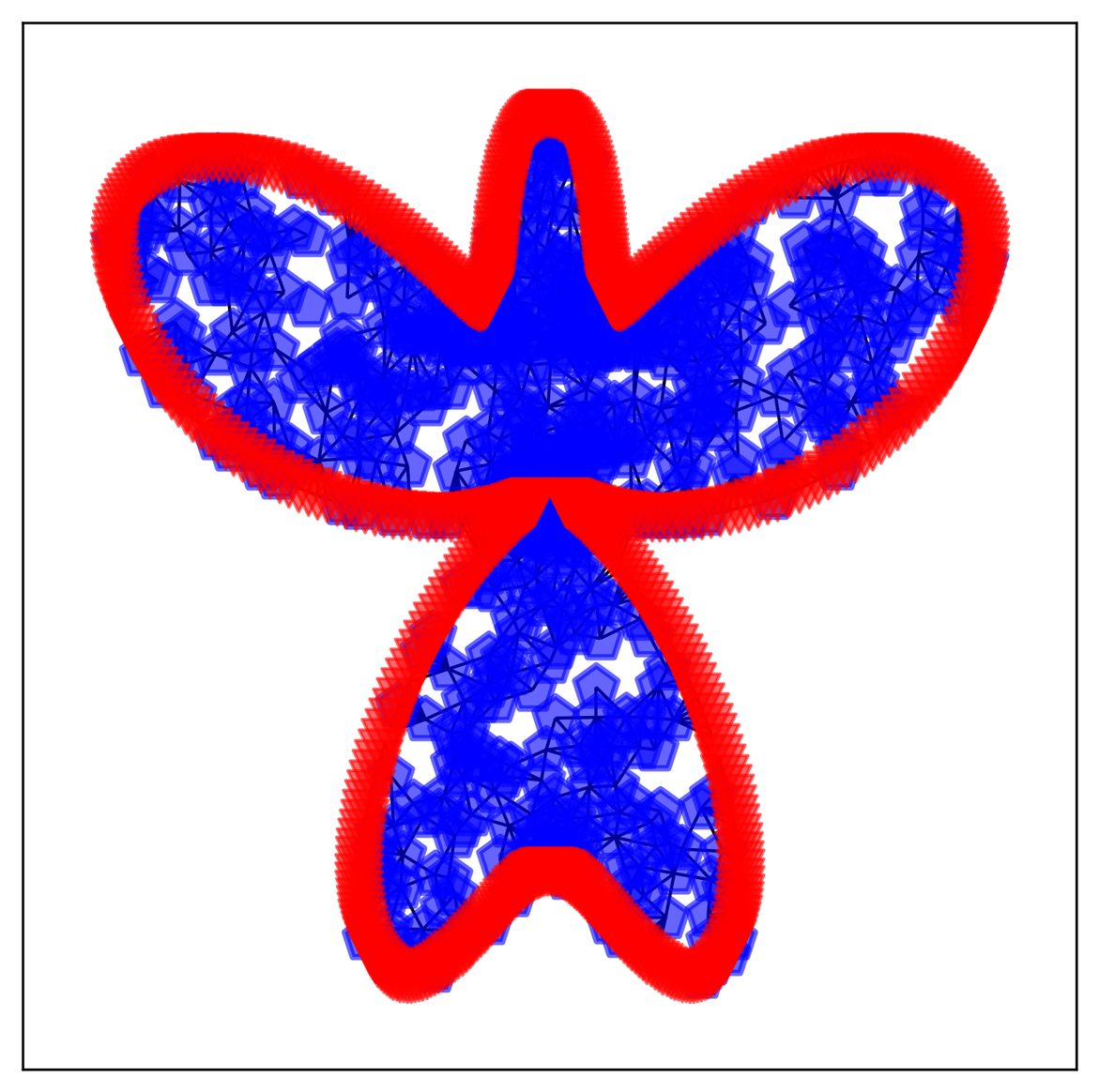}
	}
	\caption{The Delaunay algorithm is used to discrete the irregular computational domain. The graph $G = (V, E)$ has boundary nodes (red triangles) and interior nodes (blue pentagons).}
	\label{butterfly}
\end{figure}

The Delaunay algorithm is used to discrete the irregular computational domain. As a result, the graph has collocation points and boundary nodes shown in Fig.\ref{butterfly}. We would compare the predicted solution of RBF-MGN and the analytical reference at $t \in[0, T], T=2$ with the time step $\tau = 0.01s$. For the RBF-FD method, we use ph3 RBF and sample $m = 15$ nearest neighbor nodes. Gaussian RBF The RBF-FD method approximates the differential operator in local irregular space with the number of local nearest neighbor nodes $m=15$. 

\begin{figure}[htbp]
	\centering
	\subfigure[PDE residual]{
		\includegraphics[scale=0.25]{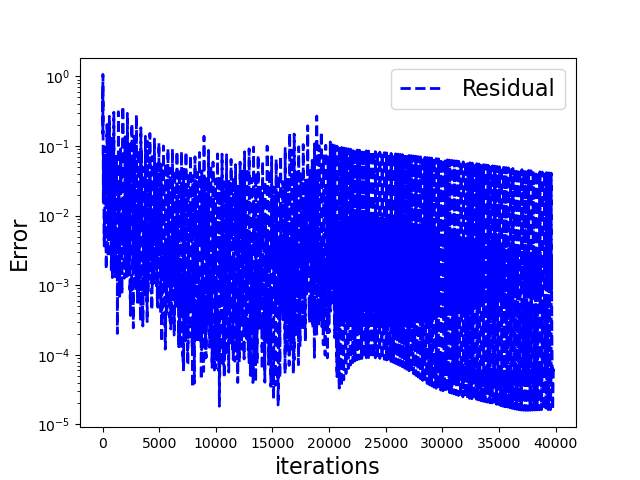}
	}%
	\subfigure[Test error]{
		\includegraphics[scale=0.25]{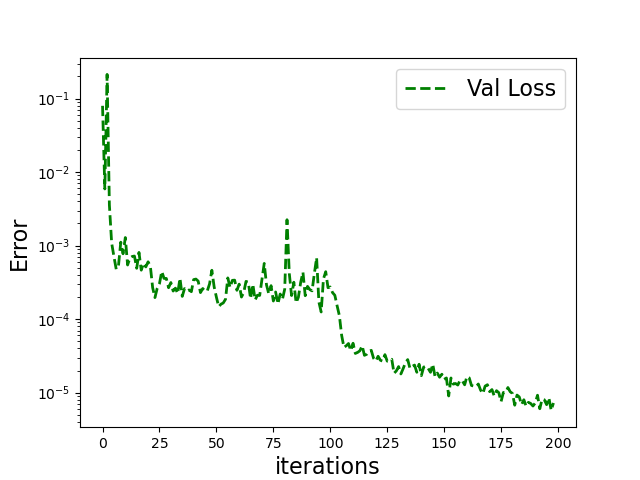}
	}%
	\centering
	\caption{Two-dimensional Poisson’s equation: The results of the PDE residual (a) and test error (b) on different spatial areas along with training iterations.}
	\label{butterfly-loss}
\end{figure}

\begin{figure}[htbp]
	\centering
	\subfigure{
		\includegraphics[scale=0.1]{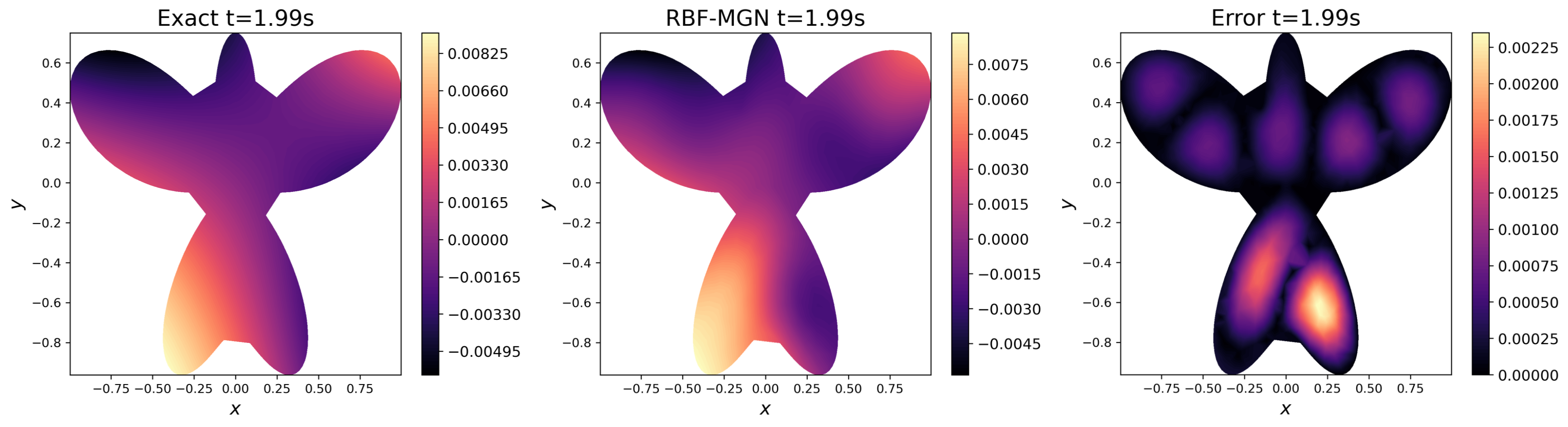}
	}
	\caption{The reuslts of two-dimensional Poisson’s equation on butterfly domain at $t = 1.99$. The predicted results are compared with the exact solutions and the difference is also presented.}
	\label{butterfly-error}
\end{figure}

\begin{figure}[htbp]
	\centering
	\subfigure{
		\includegraphics[scale=0.40]{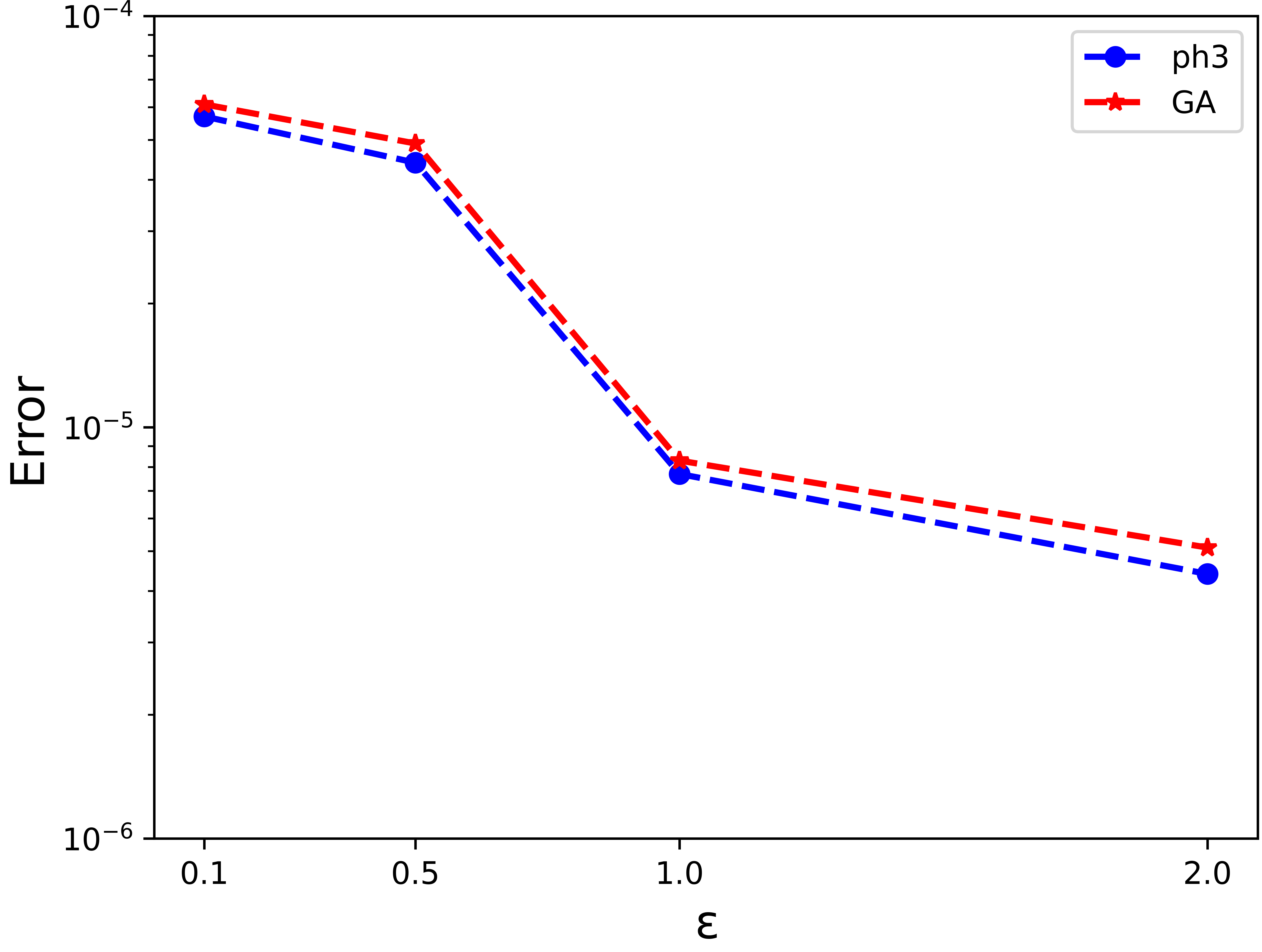}
	}
	\caption{The errors of the heat transfer problem on the butterfly domain with different $\varepsilon$.}
	\label{butterfly-weight}
\end{figure}

Setting ph3 RBF with $\varepsilon = 1.0$, we can see the RBF-MGN forward solution is almost identical to the analytical reference, and the relative prediction error is only 0.00225 according to the Fig.\ref{butterfly-loss}. In addition, this test case in Fig.\ref{butterfly-error} demonstrates that the graph-based discrete model can easily handle irregular domains with unstructured meshes, and the RBF-FD-based PDE residual can ensure that the boundary condition is also strictly satisfied. In addition, we also choose Gaussian RBF or ph3 RBF with different shape parameters $\varepsilon$ for the experiment shown in Table \ref{tab4}. It can be easily seen from Fig.\ref{butterfly-weight} that different RBF parameters affect the solution accuracy, but both give good results.

\begin{table} \fontsize{8}{8}\centering \caption{The errors of the heat transfer problem on the butterfly domain with different $\varepsilon$.}\begin{tabular}{ | c | c | c | c | c | }\hline \diagbox{RBF}{$\varepsilon$} & 0.1 & 0.5 & 1.0 & 2.0 \\\hline\hline ph3 & {5.7e-5} & {4.4e-5} & {7.7e-6} &{4.4e-6} \\\hline\hline GA & {6.1e-5} & {4.9e-5} & {8.3e-6} &{5.1e-6} \\\hline\end{tabular}\vspace{0cm}\label{tab4}\end{table}

\subsection{Two-dimensional wave equation on L-shaped domain}
In this example, we consider the two-dimensional wave equation on an L-shaped domain with free
boundary conditions.
\begin{equation}
\begin{aligned}
\frac{\partial^2 u}{\partial t^2}&=D\left(\frac{\partial^2 u}{\partial x^2}+\frac{\partial^2 u}{\partial y^2}\right),  \quad t \in[0, T] \\ (x, y) \in \Omega &= Polygon[[0, 0], [2, 0], [2, 1], [1, 2], [0, 2], [1, 1]].
\end{aligned}
\end{equation}

\begin{figure}[htbp]
	\centering
	\subfigure{
		\includegraphics[scale=0.5]{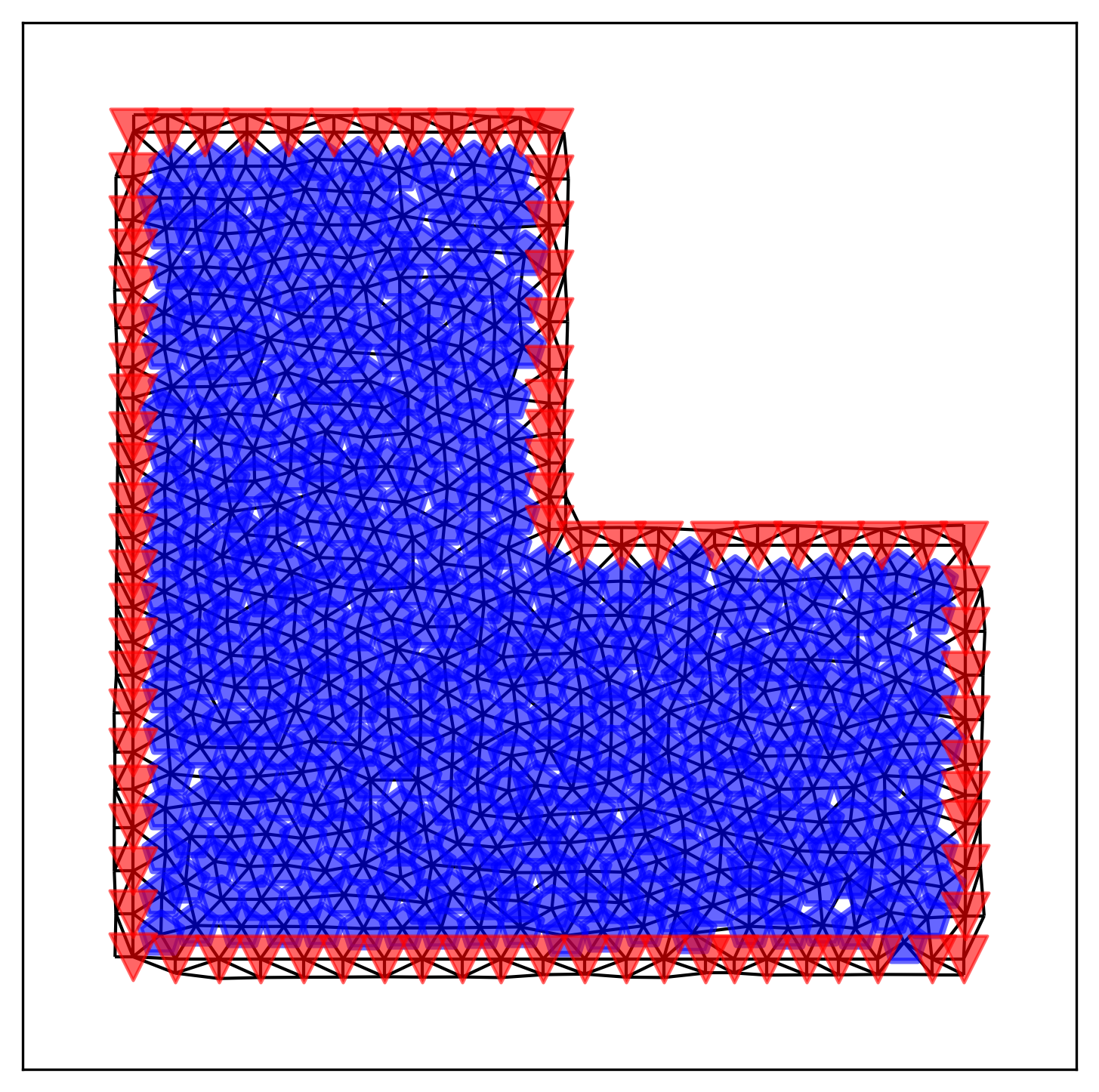}
	}
	\caption{The Delaunay algorithm is used to discrete the L-shaped domain. The graph $G = (V, E)$ has boundary nodes (red triangles) and interior nodes (blue pentagons).}
	\label{L}
\end{figure}

\begin{figure}[htbp]
	\centering
	\subfigure{
		\includegraphics[scale=0.1]{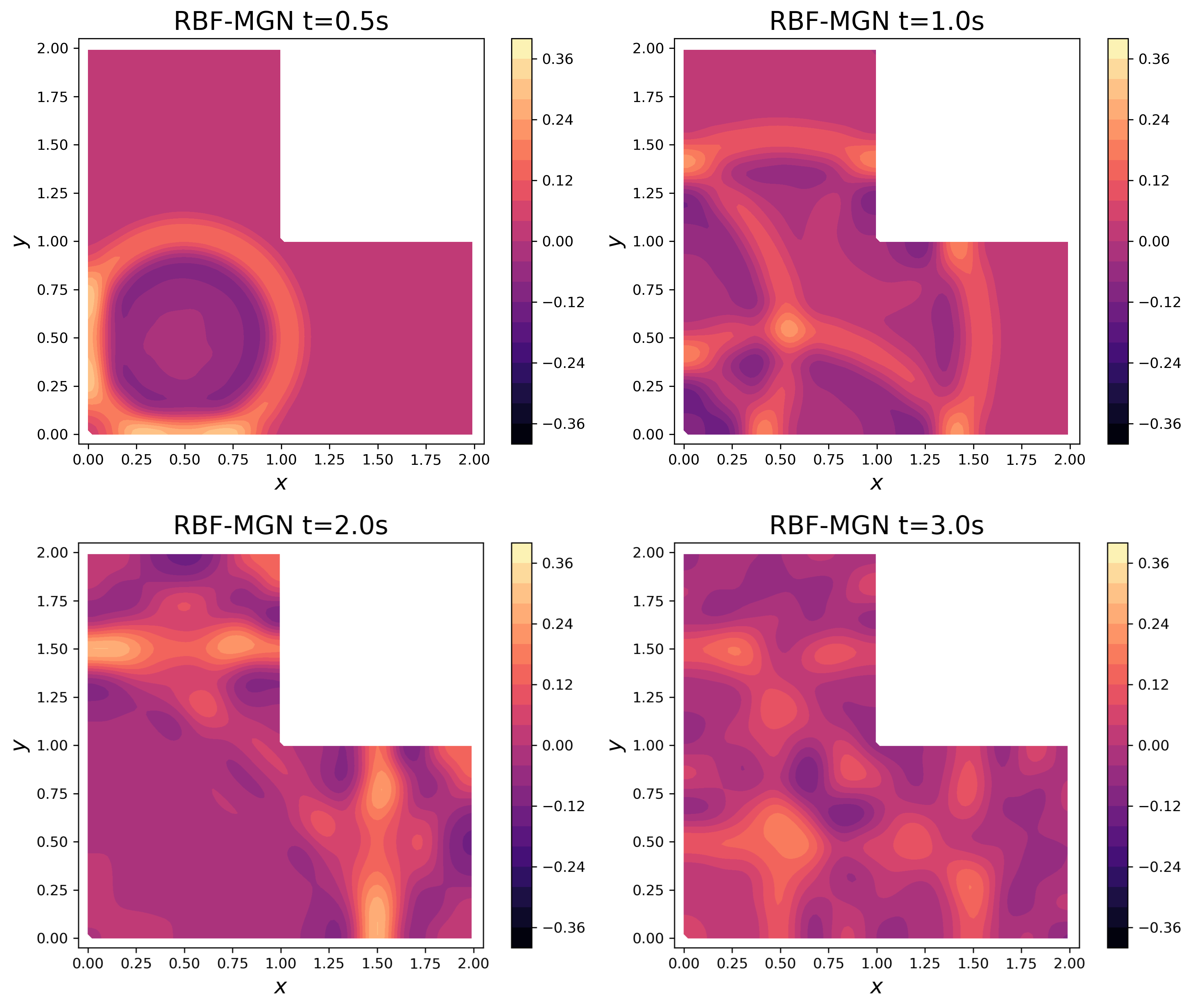}
	}
	\caption{The reuslts of two-dimensional wave equation on L-shaped domain at different time steps.}
	\label{L-error}
\end{figure}

We set the initial displacements in this problem at the interior and boundary. The velocities are zero, and the PDE parameter is $1e-6$. Triangulate the irregular region as shown in the Fig.\ref{L} to represent the graph $G=(V, E)$, 405 observations are randomly sampled in the domain, including 84 boundary nodes. We aim to attain the solution at $t \in[0, T]$ with the time step $\tau = 0.1s$ through the proposed method. The Adam optimizer is applied to update the neural network parameters in the iterations. We also use ph3 RBF and sample $m = 25$ nearest neighbor nodes. RBF-MGN could accurately recover the temperature at $t = 1.99$ as shown in Fig.\ref{L-error}. We show the predicted solutions of RBF-MGN at different time instants $T = 0.50, 1.00, 2.00, 3.00$.

\begin{figure}[htbp]
 	\centering
 	\subfigure{
 		\includegraphics[scale=0.35]{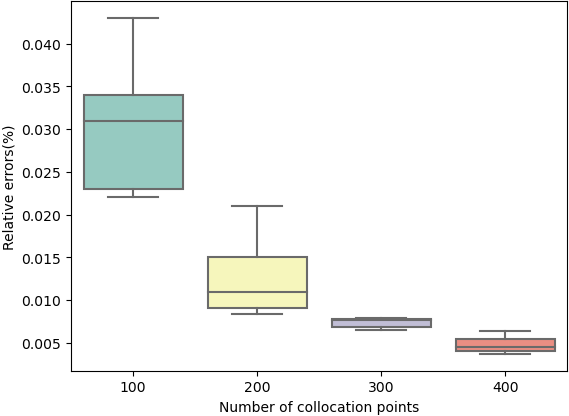}
 	}
   \subfigure{
   	\includegraphics[scale=0.35]{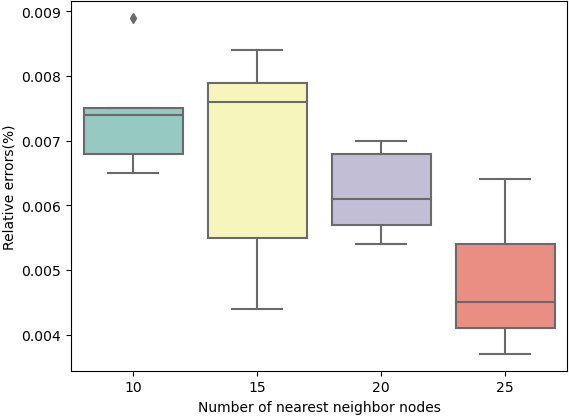}
   }%
 	\caption{Two-dimensional wave equation on L-shaped domain: Boxplot of the relative errors ($\%$) with different numbers of collocation points (Left) and nearest neighbor nodes (Right).}
 	\label{L-pa}
\end{figure}

\begin{figure}[htbp]
	\centering
	\subfigure{
		\includegraphics[scale=0.40]{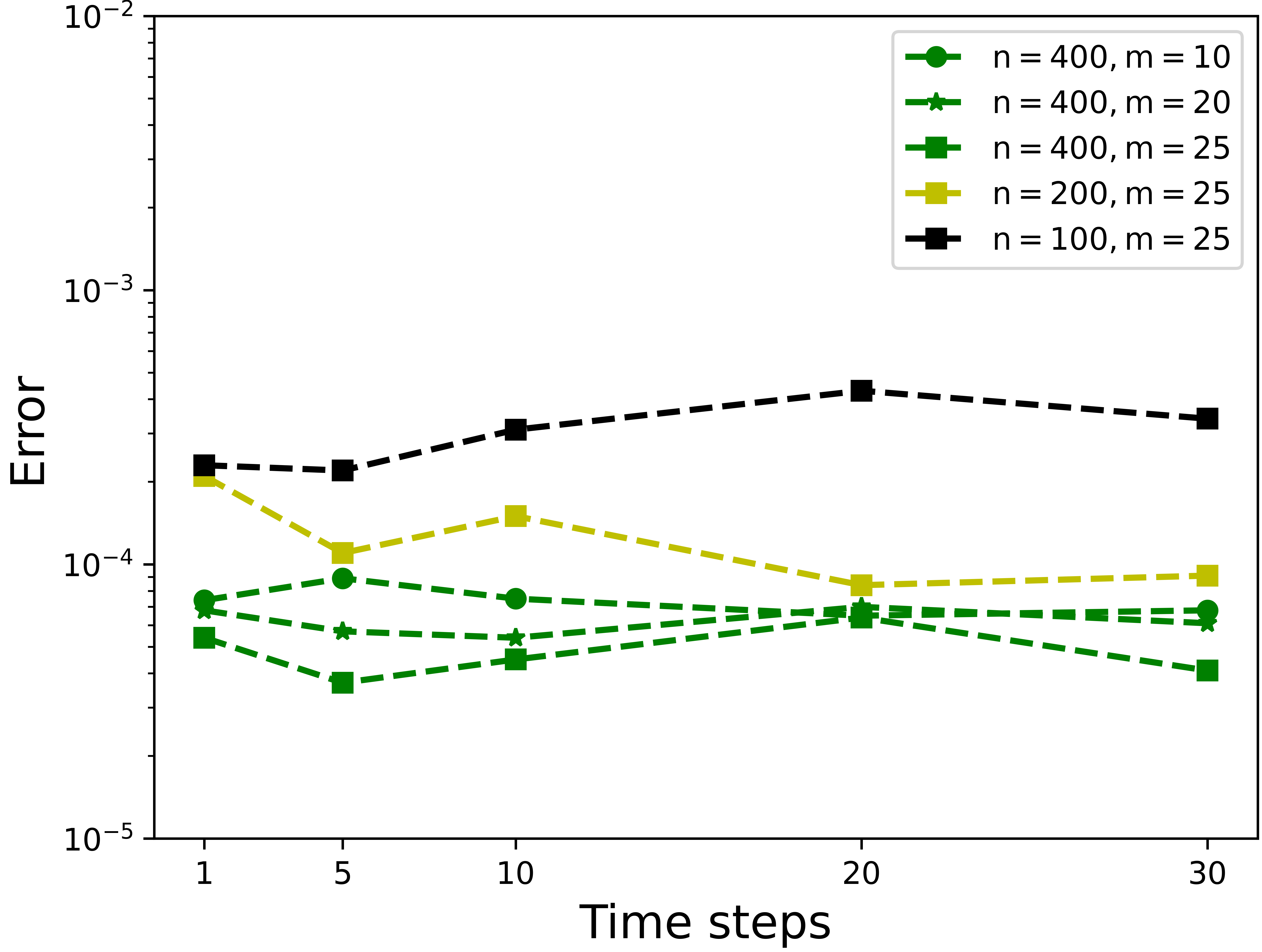}
	}
	\caption{The errors of the heat transfer problem on the butterfly domain with different parameters $n$ and $m$.}
	\label{L-weight}
\end{figure}

To further scrutinize the performance of the proposed method, we have performed a systematic study concerning the size of the observation dataset $n$. First, 100, 200, 300, and 400 points in the computational domain are randomly selected to represent the graph. Respectively. Table \ref{tab5} shows the results at different time steps, where the number of nearest neighbor nodes is fixed to  $m = 25$. Second, Setting $n = 400$, we also analyze the performance of RBF-MGN in different numbers of nearest neighbor nodes $m$ in Table \ref{tab6}. As shown in Fig.\ref{L-pa}, RBF-MGN is capable of achieving a more accurate solution when trained with the most miniature set of collocation points. RBF-MGN is also insensitive to this parameter $m$ (10, 15, 20, 25), which also indicates that the powerful approximation capability of the neural network can ignore the effect of the parameter on the RBF-FD approximate differential operator. As is shown in Fig.\ref{L-weight}, for all cases, RBF-MGN could achieve acceptable minor errors, especially with $n=400, m = 25$. 

\begin{table} \fontsize{8}{8}\centering \caption{The relative errors ($\%$) at different time steps of the Two-dimensional wave equation on L-shaped domain with numbers of collocation points $n$.}\begin{tabular}{ | c | c | c | c | c | c | }\hline \diagbox{$n$}{$setps$} & 1 & 5 & 10 & 20 & 30 \\\hline\hline 100 & {0.023} & {0.022} & {0.031} &{0.043} &{0.034} \\\hline\hline 200 & {0.021} & {0.011} & {0.015} &{0.0084} &{0.0091} \\\hline\hline 300 & {0.0078} & {0.0079} & {0.0068} &{0.0077} & {0.0065}\\\hline\hline 400 & {0.0054} & {0.0037} & {0.0045} &{0.0064} & {0.0041}\\\hline\end{tabular}\vspace{0cm}\label{tab5}\end{table}
 
\begin{table} \fontsize{8}{8}\centering \caption{The relative errors ($\%$) at different time steps of the Two-dimensional wave equation on L-shaped domain with numbers of nearest neighbor nodes $m$.}\begin{tabular}{ | c | c | c | c | c | c | }\hline \diagbox{$n$}{$setps$} & 1 & 5 & 10 & 20 & 30 \\\hline\hline 10 & {0.0074} & {0.0089} & {0.0075} &{0.0065} &{0.0068} \\\hline\hline 15 & {0.0044} & {0.0076} & {0.0079} &{0.0084} &{0.0055} \\\hline\hline 20 & {0.0068} & {0.0057} & {0.0054} &{0.0070} & {0.0061}\\\hline\hline 25 & {0.0054} & {0.0037} & {0.0045} &{0.0064} & {0.0041}\\\hline\end{tabular}\vspace{0cm}\label{tab6}\end{table}

\section{Conclusions}
This paper proposes a physics-informed framework (RBF-MGN) based on GNNs and RBF-FD to solve spatio-temporal PDEs. The GNNs and RBF-FD are introduced into physics-informed learning to handle irregular domains with unstructured meshes better. Combined with the boundary conditions, a high-precision difference format of the differential equations is constructed to guide model training. The numerical results from several Poisson’s equations on complex domains have shown the effectiveness of the proposed method. Furthermore, we also tested the robustness of the RBF-MGN with different time steps, PDE parameters, different numbers of collocation points, and several types of RBFs. 

It should be noted that there are fluctuations in the loss functions constructed based on RBF-FD, and we should further infer the reasons for this phenomenon and effectively reduce the gradient fluctuations of these loss functions.
\bibliographystyle{model1-num-names}
\bibliography{reference}

%

\end{document}